%% file: aaai22.tex
\relax
\documentclass[letterpaper]{article} 
\pdfoutput=1
\usepackage{aaai22}  
\usepackage{times}  
\usepackage{helvet}  
\usepackage{courier}  
\usepackage[hyphens]{url}  
\usepackage{graphicx} 
\usepackage{natbib}  
\usepackage{caption} 
\DeclareCaptionStyle{ruled}{labelfont=normalfont,labelsep=colon,strut=off} 
\usepackage{algorithm}
\usepackage{algorithmic}
\usepackage{color}

\usepackage{newfloat}
\usepackage{listings}
\usepackage{multirow}
\usepackage{amsmath}
\usepackage{amssymb}
\usepackage{float}

\floatstyle{ruled}
\newfloat{listing}{tb}{lst}{}
\floatname{listing}{Listing}
\title{Contextual Modeling for 3D Dense Captioning on Point Clouds}
\author {
    Yufeng Zhong\textsuperscript{\rm 1,2}, 
    Long Xu\textsuperscript{\rm 1}$^*$, 
    Jiebo Luo\textsuperscript{\rm 3}, 
    Lin Ma\textsuperscript{\rm 4}\thanks{Corresponding authors.}
}
\affiliations {
    \textsuperscript{\rm 1} National Space Science Center, Chinese Academy of Sciences\\
    \textsuperscript{\rm 2} University of Chinese Academy of Sciences\\
     \textsuperscript{\rm 3} University of Rochester\\ 
     \textsuperscript{\rm 4} Meituan\\
    zhongyufeng21@mails.ucas.ac.cn, lxu@nao.cas.cn, jluo@cs.rochester.edu, forest.linma@gmail.com
}

\usepackage{bibentry}

\begin{document}

\maketitle

\begin{abstract}
3D dense captioning, as an emerging vision-language task, aims to identify and locate each object from a set of point clouds and generate a distinctive natural language sentence for describing each located object.
However, the existing methods mainly focus on mining inter-object relationship, while ignoring contextual information, especially the non-object details and background environment within the point clouds, thus leading to low-quality descriptions, such as inaccurate relative position information.
In this paper, we make the first attempt to utilize the point clouds clustering features as the contextual information to supply the non-object details and background environment of the point clouds and incorporate them into the 3D dense captioning task. We propose two separate modules, namely the Global Context Modeling (GCM) and Local Context Modeling (LCM), in a coarse-to-fine manner to perform the contextual modeling of the point clouds. Specifically, the GCM module captures the inter-object relationship among all objects with global contextual information to obtain more complete scene information of the whole point clouds. The LCM module exploits the influence of the neighboring objects of the target object and local contextual information to  enrich the object representations.
With such global and local contextual modeling strategies, our proposed model can effectively characterize the object representations and contextual information and thereby generate comprehensive and detailed descriptions of the located objects. 
Extensive experiments on the ScanRefer and Nr3D datasets demonstrate that our proposed method sets a new record on the 3D dense captioning task, and verify the effectiveness of our raised contextual modeling of point clouds.
\end{abstract}

\input{1_introduction}

\input{2_relatedwork}
\input{3_method}
\input{4_experiment}
\input{5_coclusion}



\section{Reference}
\label{sec:reference_examples}

\nobibliography*

\bibentry{xu2015show}.\\[.2em]
\bibentry{lu2017knowing}.\\[.2em]
\bibentry{zhang2021rstnet}.\\[.2em]
\bibentry{anderson2018bottom}.\\[.2em]
\bibentry{kim2019dense}.\\[.2em]
\bibentry{yin2019context}.\\[.2em]
\bibentry{li2019entangled}.\\[.2em]
\bibentry{wang2019hierarchical}.\\[.2em]
\bibentry{luo2021dual}.\\[.2em]
\bibentry{johnson2016densecap}.\\[.2em]
\bibentry{yang2017dense}.\\[.2em]
\bibentry{chen2021scan2cap}.\\[.2em]
\bibentry{wang2022spatiality}.\\[.2em]
\bibentry{song2019much}.\\[.2em]
\bibentry{jiao2022more}.\\[.2em]
\bibentry{chen2020scanrefer}.\\[.2em]
\bibentry{achlioptas2020referit3d}.\\[.2em]
\bibentry{yuan2022x}.\\[.2em]
\bibentry{chen2021d3net}.\\[.2em]
\bibentry{cai20223djcg}.\\[.2em]
\bibentry{qi2019deep}.\\[.2em]
\bibentry{liu2021group}.\\[.2em]
\bibentry{qi2017pointnet++}.\\[.2em]
\bibentry{vaswani2017attention}.\\[.2em]
\bibentry{rennie2017self}.\\[.2em]
\bibentry{paszke2016enet}.\\[.2em]
\bibentry{deshpande2019fast}.\\[.2em]
\bibentry{pan2020x}.\\[.2em]
\bibentry{datta2019align2ground}.\\[.2em]
\bibentry{zhong2020comprehensive}.\\[.2em]

\nobibliography{aaai22}

\end{document}

%% file: 1_introduction.tex
\section{Introduction}

\begin{figure}[t]
\centering
\includegraphics[width=\linewidth]{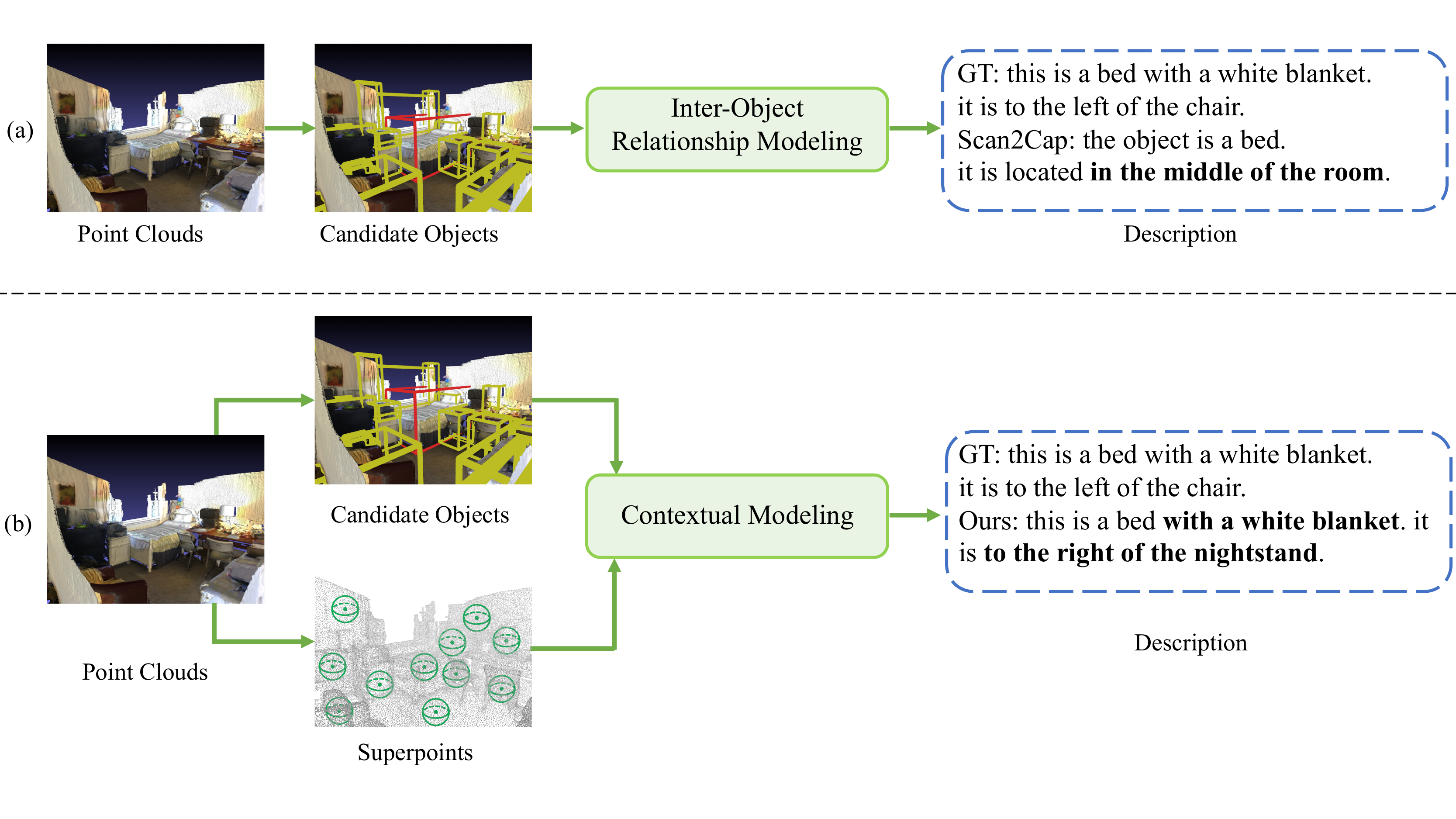} 
\vspace{-0.8cm}
\caption{
Different pipelines for 3D dense captioning. 
(a) The traditional pipeline detects the candidate objects and performs the inter-object relationship modeling to generate the corresponding caption.  
(b) Our proposed method  not only detects the candidate objects but also yields the superpoints depicting the contextual information, and thereafter performs the contextual modeling to generate the resulting caption. }
\vspace{-0.5cm}
\label{fig1}
\end{figure}

With continuous advance in deep learning, there has been increasing attentions on the intersection of computer vision (CV) and natural language processing (NLP), such as image captioning~\cite{xu2015show, lu2017knowing, zhang2021rstnet} and dense image captioning~\cite{johnson2016densecap, yang2017dense, kim2019dense}.
Image captioning aims at generating a single sentence for an image, while dense image captioning is the task of localizing multiple objects in a given image and describing each object by natural language sentences.
Such captioning tasks are mainly restricted to 2D images.
Recently, some researchers have extended the dense captioning task from 2D images to 3D point clouds and proposed the 3D dense captioning task~\cite{chen2021scan2cap, jiao2022more, wang2022spatiality}.
Different from 2D images captured from only one single viewpoint, 3D point clouds~\cite{chen2020scanrefer, achlioptas2020referit3d} can provide complete physical information of objects, such as the actual size and space locations in the real environment. 
However, the complexity of 3D point clouds makes the 3D dense captioning task even more challenging.

In general, the existing 3D dense captioning methods~\cite{chen2021scan2cap, jiao2022more, wang2022spatiality} usually adopt a detector to generate candidate objects and model the inter-object relationship to learn the complicated object features for generating the corresponding descriptions, as illustrated in Fig.\ref{fig1}~(a).
Concretely,~\citet{chen2021scan2cap} propose the first baseline method, namely Scan2Cap, which depicts the relationship  between the target object and candidate objects by establishing a message-passing network.
\citet{jiao2022more} propose to progressively encode inter-object relations among candidate objects by a multi-order relation mining model.
\citet{wang2022spatiality} design a spatiality-guided encoder to investigate the relative spatiality between the candidate objects.
However, existing methods only focus on mining the relationships between candidate objects, while ignoring the contextual information within the point clouds.
Unfortunately, the lack of contextual information, especially the non-object details and background environment within the whole point clouds, may result in low-quality and inaccurate descriptions.
For example, as shown in Fig.\ref{fig1}~(a), Scan2Cap generates wrong object position ``\texttt{\small{in the middle of the room}}'', with only modeling the inter-object relationship between candidate objects.


To overcome the shortcomings of existing approaches, we make the first attempt to incorporate the contextual information into the 3D dense captioning task, with the proposed pipeline illustrated in Fig.\ref{fig1}~(b). 
Specifically, taking the input point clouds, a detector is used to extract candidate objects and superpoints.
Here, the superpoints are obtained by sampling and clustering the point clouds step by step, which distributes evenly not only within the interior but also outside of objects. As such, they can enrich the candidate object representations and meanwhile capture the non-object details and background environment of the whole point clouds.
Based on the obtained candidate objects and superpoints, we perform the contextual modeling globally and locally, which is realized in two individual modules, namely Global Context Modeling (GCM) and Local Context Modeling (LCM).
GCM exploits the relationships among all the candidate objects and superpoints to obtain more complete scene information of the whole point clouds.
LCM relies on the neighboring objects and superpoints of the target object to attain more complicated object features, which can further enrich the object representation.
By fusing the exploited global and local features, our proposed model is able to generate more comprehensive and detailed descriptions. For example, as illustrated in Fig.\ref{fig1}~(b), by considering the candidate objects and superpoints, our model can recognize the ``\texttt{\small{white blanket}}'' on the bed and infer the correct relative position of the bed and nightstand.
We conduct experiments on the 3D dense captioning benchmark datasets ScanRefer and Nr3D to quantitatively verify the effectiveness of our model. And the experiment results show that  our proposed model achieves  the state-of-the-art performance.

In summary, the contributions of this paper lie in the following three-fold: 
\begin{enumerate}
\item We make the first attempt to incorporate the contextual information of the point clouds into 3D dense captioning task.
    
\item We design two modules, namely Global Context Modeling (GCM)  and Lobal Context Modeling (LCM), to effectively mine the relationship between the target object and the objects as well as the superpoints. Specifically, GCM exploits the relationships between all objects and superpoints, and LCM establishes the relationship between the target object and its neighboring objects as well as superpoints.

\item Extensive experiments demonstrate that our proposed method achieves the state-of-the-art performances on ScanRefer (\textbf{54.30\% CiDEr@0.5IoU}) and Nr3D (\textbf{37.37\% CiDEr@0.5IoU}) datasets.

\end{enumerate}




%% file: 2_relatedwork.tex
\section{Related Work}

\subsection{Image Captioning}
In the last few years, image captioning has attracted significant research interest and many researchers have proposed a series of methods~\cite{xu2015show, anderson2018bottom, zhang2021rstnet}.
Some methods~\cite{xu2015show, lu2017knowing, deshpande2019fast} utilize the visual representation of multiple image regions as input to generate descriptions.
Following BUTD~\cite{anderson2018bottom}, a trend is to combine the attention mechanism and object detection to improve the performance of the image captioning~\cite{datta2019align2ground, zhong2020comprehensive}.
After that, with the thriving of the Transformer~\cite{vaswani2017attention}, Transformer-based methods~\cite{li2019entangled, pan2020x, zhang2021rstnet} have become predominant in the field of image captioning.


\subsection{Dense Captioning}
The dense captioning task is introduced by~\citet{johnson2016densecap}, as a variant of image captioning that tries to localize and describe all objects in images. 
\citet{yang2017dense} propose a context fusion component to emphasize the visual cues of the surrounding salient image regions as context features to generate better descriptions.
\citet{kim2019dense} capture the relational information between detected objects, which provide exact concepts and richer representation.
Further,~\citet{yin2019context} integrate mutual interactions from neighboring contextual regions of the target object and global visual information for more accurate dense captioning.
\citet{song2019much} propose a semantically symmetric LSTM model, which not only understands particular regions of the image but also utilizes global features as contextual information.

\subsection{3D Dense Captioning}


3D dense captioning is closely related to the dense captioning task.
In the 3D dense captioning task, \citet{chen2021scan2cap} 
propose the first baseline model Scan2Cap. 
Scan2Cap uses a message-passing network to mine the relationship between the target object and different objects, and enhance object relation representation.
\citet{jiao2022more} propose a Multi-Order RElation mining model to encode object relations progressively in the graph.
SpaCap3D~\cite{wang2022spatiality} uses a spatiality-guided encoder and an object-centric decoder to learn the contribution of surrounding objects to the target object.
Despite achieving impressive results, these methods ignore the contextual information and only consider the influence of the adjacent objects of the target object. 
Therefore, to make full use of the contextual information, we develop a contextual modeling method that integrates global and local contextual information of the point clouds.

\begin{figure*}[htb]
\centering
\includegraphics[width=0.96\linewidth]{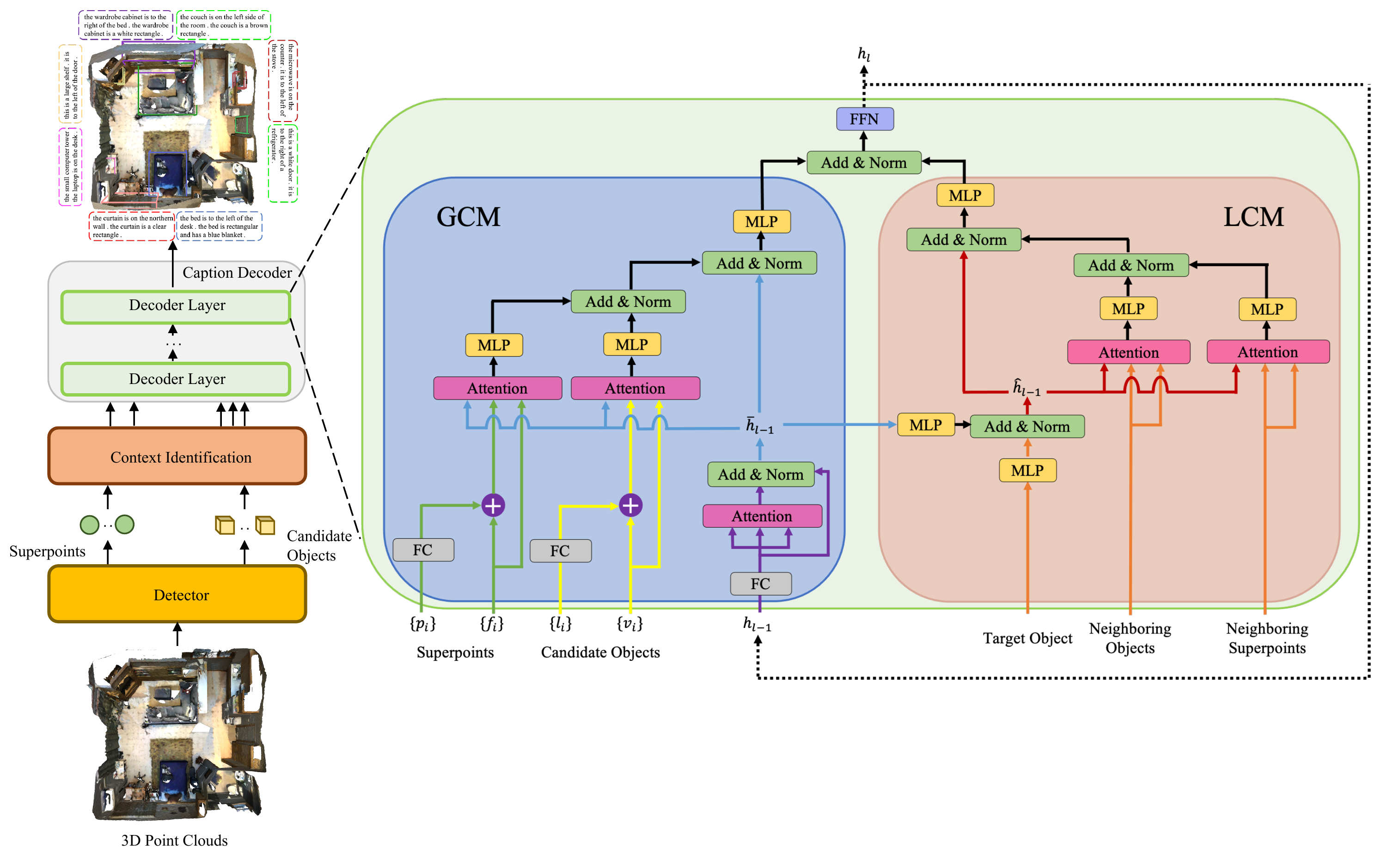} 
\caption{An overview of our proposed method, which consists of three components: a detector, a context identification module, and a caption decoder. 
Given a set of point clouds, the detector extracts the superpoints and candidate objects.
For all superpoints and candidate objects, the context identification module specifies the target object and selects the neighboring objects and superpoints around the target object.
Each layer in our caption decoder consists of two modules, namely GCM and LCM.
GCM learns the global features from all objects and superpoints to obtain more complete scene information of the whole point clouds. LCM yields the local features from the neighboring objects and superpoints to enrich each object representation. 
Finally, the global and local features are fused together to generate the corresponding descriptions.
}
\vspace{-0.2cm}
\label{fig2}
\end{figure*}

%% file: 3_method.tex
\section{Method}
\subsection{Framework Overview}
Fig.~\ref{fig2} illustrates our proposed framework by incorporating the contextual information for 3D dense captioning, which consists of three modules: a detector, a context identification module, and a caption decoder.
First, the detector takes the point clouds as input and generates the superpoints and candidate objects, which are regarded as the contextual information and object features, respectively. 
Afterwards, to tackle the field-of-view issue, it is also necessary to obtain more distinctive information about the target object.
Hence, the context identification module is used to specify the target object and selects the objects and superpoints nearest to the target object as its neighboring objects and superpoints.
{Each layer} in our caption decoder consists of two modules, namely GCM and LCM. GCM learns the global features from all objects and superpoints to obtain more complete scene information of the whole point clouds. LCM yields the local features from the neighboring objects and superpoints to enrich each object representations.
Finally, we combine the global and local features to generate comprehensive and detailed descriptions.

\subsection{Detector}
Given a set of input point clouds, the purpose of the detector is to generate superpoints and candidate objects.
To obtain effective feature representations from the original disordered and scattered point clouds, we first sample and cluster the point clouds step by step to generate the superpoints (the procedure is similar to most point clouds backbone, such as PointNet++~\cite{qi2017pointnet++}).
Here, we denote the superpoints as $\left\{\left(p_{i}, f_{i}\right)\right\}_{i=1}^{N_{\text{SP}}}$, where $N_{\text{SP}}$ is the number of superpoints, $p_{i}$ is the center coordinate $(x,y,z)$ and $f_{i}$ represents the visual feature vector of the corresponding superpoint.
Please note that the superpoints distributes evenly not only within the interior but also outside of objects. As such, they can enrich the candidate object representations and meanwhile capture the non-object details and background environment of the whole point clouds.
Besides, 
the  superpoints can be further aggregated to extract candidate objects accompanied with their corresponding bounding box.
We denote the candidate objects as $\left\{\left(l_{i}, v_{i}\right)\right\}_{i=1}^{{{N}_\text{{OBJ}}}}$, where $N_{\text{OBJ}}$ is the number of candidate objects, $l_{i}$ is the bounding box and $v_{i}$ represents the appearance feature vector of the candidate object. 



\subsection{Context Identification}

\begin{figure}
\centering
\includegraphics[width=\linewidth]{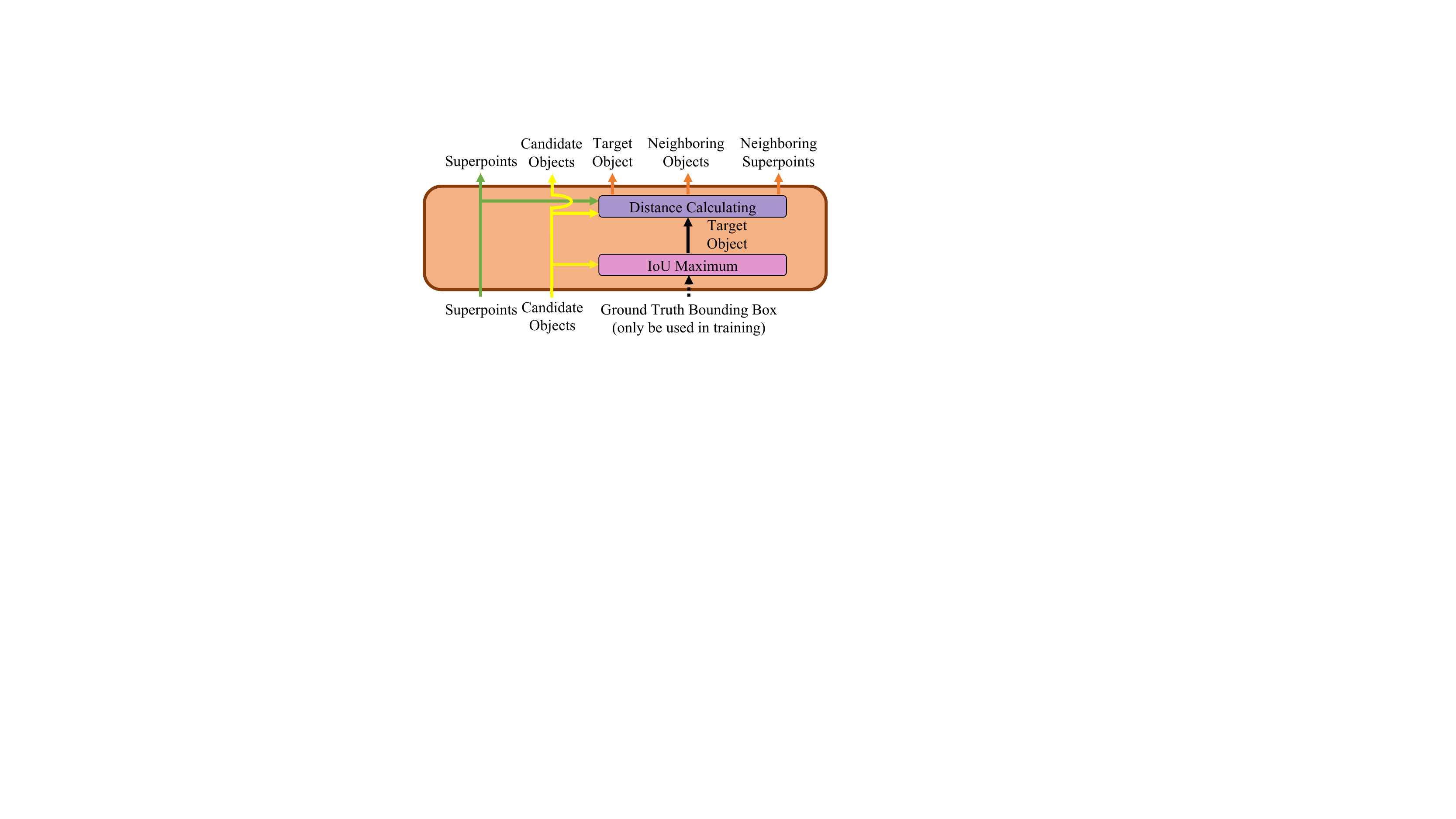} 
\caption{The context identification module specifies the target object and selects the neighboring objects and superpoints close to the target object as its context.}
\vspace{-0.3cm}
\label{fig3}
\end{figure}

For the 3D dense captioning task, we need to specify the target object from all the candidate objects and identify its context for decoding the corresponding caption during the training and inference stage.
As shown in Fig.~\ref{fig3}, during training, we compare all the predicted bounding box $l_{i}$ with the ground truth bounding box of the object to be described, and select the object with the highest Intersection over Union (IoU) score as the target object.
In addition, to solve the field-of-view problem, it is necessary to mine the spatial relationship of the target object and its neighboring objects and superpoints.
One on hand, we calculate the distance between all objects and the target object and then select the neighboring objects close to the target object.
On the other hand, referring to the distances between the target object and the superpoints, we also select the neighboring superpoints close to the target object, which can further enrich the object representations and express the non-object details and background environment. During the inference, we take all objects generated by the detector as the target object one by one without any ground truth bounding box and thereafter identify its context.

\subsection{Caption Decoder}

Our proposed caption decoder consists of multiple decoder layers, stacking together to characterize each located candidate object of the point clouds and thereby yield its corresponding caption. Each decoder layer consists of two separate modules, namely GCM and LCM. 
On one hand, to obtain more comprehensive understanding of the whole point clouds,  GCM takes all the candidate objects and superpoints extracted from the detector as input to generate the global representations.
On the other hand, to enrich the object representations,  LCM yields the local representations by exploiting the relationships between the target object and the neighboring objects as well as superpoints. 
Finally, the global and local representations are fused together to generate the corresponding descriptions.

\subsubsection{Global Context Modeling.}
To obtain more comprehensive understanding of the whole point clouds, GCM takes all the candidate objects and superpoints into account for yielding the global representations. 
The superpoints distribute evenly not only within the interior but also outside of objects. As such, the superpoints together with candidate objects can well characterize the whole point clouds, for example, identifying  the  relative position of each located object.
The detailed pipeline of GCM is illustrated  in Fig.~\ref{fig2}.

For the $l$-th layer of caption decoder, we feed the fused feature $h_{l-1}$ yielded from the previous decoder layer into an FC layer. Afterwards, an Attention layer with an Add \& Norm layer is performed to generate the feature $\overline{h}_{l-1}$.
Considering that the bounding boxes $\left\{ l_{i} \right\}$ contains the position information of candidate objects, we take the bounding boxes $\left\{ l_{i} \right\}$ into an FC layer and add the appearance features $\left\{ v_{i} \right\}$ of candidate objects to get object embeddings.
After that, we feed the feature $\overline{h}_{l-1}$, object embeddings, and appearance features $\left\{ v_{i} \right\}$ to an Attention layer with an MLP layer to generate the enhanced object features, which aggregates the fused feature from the previous decoder layer and the initial object features.
Similarly, we feed the center coordinates $\left\{ p_{i} \right\}$ into an FC layer and add the visual features $\left\{ f_{i} \right\}$ of superpoints to obtain superpoint embeddings.
Next, we take the feature $\overline{h}_{l-1}$, superpoint embeddings, and visual features $\left\{ f_{i} \right\}$ to an Attention layer with an MLP layer to generate the enhanced superpoint features, which aggregates the fused feature from the previous decoder layer and the initial superpoint features.
Then, we use the enhanced object features and enhanced superpoint features as the input of an Add \& Norm layer to get the fused visual features.
Finally, we integrate the feature $\overline{h}_{l-1}$ and the fused visual features by using an Add \& Norm layer with an MLP layer to obtain the global representations.
Please note that we use the previously generated word as $h_{0}$ in the first decoder layer.

\subsubsection{Local Context Modeling.}
To generate the comprehensive and detailed descriptions, each located candidate objects needs to be more accurately depicted and characterized. Besides the category information of the object, the attribute information of the object, such as color and accessories, can further help improving the quality of the caption. However, these attribute information seem to be  difficult to be captured by only referring the candidate objects as well as their neighboring objects. We propose LCM  to further incorporate  the neighboring superpoints of the target object, which contain detailed local features of the target object and its surrounding background information, to enrich the representation of the target object. The detailed pipeline of LCM is shown in Fig.~\ref{fig2}.

First, we feed the appearance feature of the target object and the feature $\overline{h}_{l-1}$ into an MLP layer, respectively, and then use an Add \& Norm layer to obtain the feature $\hat{h}_{l-1}$.
Next, the feature $\hat{h}_{l-1}$ and the appearance features of the neighboring objects are input to an Attention layer with an MLP layer to generate the target-oriented object features.
Besides, we take the feature $\hat{h}_{l-1}$ and visual features of the neighboring superpoints into an Attention layer with an MLP layer to output the target-oriented superpoint features.
Since the neighboring objects and superpoints of the target object already contain the relative position information, so we do not use the bounding boxes and center coordinates.
Then, we input the target-oriented object features and target-oriented superpoint features to an Add \& Norm layer to get the target-oriented visual features.
After that, we use the feature $\hat{h}_{l-1}$ and target-oriented visual features as the input of an Add \& Norm layer with an MLP layer to obtain the local representations.

{Finally, we adopt an Add \& Norm layer to fuse the global and local representations, and use an FFN layer to generate the fused feature $h_{l-1}$, which will be taken as the input of the next decoder layer. 
The output fused feature $h_{L}$ of the last decoder layer is fed into an FC layer to predict the next word.}

\subsection{Training and Inference}
During the training phase, 
to generate the descriptions, we adopt a teacher-forcing strategy that uses the ground truth captioning token as the previously generated word.
Besides, we design a multi-stage training scheme to train our model.
First, we train the caption decoder with the frozen pre-trained detector and then fine-tune the detector and decoder together.
Given the ground truth description $y_{1: T}$ of the target object {with the length of $T$} and a captioning model with parameters $\theta$, we optimize the generated word token probabilities by a conventional cross-entropy loss:
\begin{equation}
\label{eq:6}
L_{\text{XE}}(\theta)=-\sum_{t=1}^{T} \log \left(p_{\theta}\left(y_{t} \mid y_{1: t-1}\right)\right).
\end{equation}
To further boost the captioning performances, we follow the self-critical sequence training scheme described in~\cite{rennie2017self} and adopt the CiDEr score as the reward:
\begin{equation}
\label{eq:7}
L_{\text{SC}}(\theta)=-\mathbf{E}_{y_{1: T} \sim p_{\theta}}\left[r\left(y_{1: T}\right)\right].
\end{equation}

During the inference phase, our model takes a set of point clouds as input and generates the descriptions of all the detected objects. First,  the given point clouds are fed into the detector to extract all candidate objects and superpoints. And then, the context identification module takes all candidate objects as the target object one by one without any ground truth bounding box and generates their corresponding neighboring objects and superpoints. Afterwards, for each target object, we input all objects and superpoints, as well as the neighboring objects and superpoints to the caption decoder for producing the corresponding captions. 


%% file: 4_experiment.tex
\section{Experiments}

In this section, we first describe our experimental settings, including the datasets, evaluation metrics, as well as the implementation details.
Afterwards, we compare our proposed method with the state-of-the-art methods.
Finally, we present a more detailed analysis of our method, including the qualitative results and ablation studies.

\begin{table*}[!htb]
\centering
\caption{Performance comparisons of 3D dense captioning results obtained by our and the state-of-the-art methods using VoteNet as the detector on the ScanRefer dataset.  ``-'' indicats that the results have not been disclosed. ``*'' indicates training with the CiDEr score reward.}
\label{tab:1}
\begin{tabular}{l | c | c c c c c}
\hline
{Method} & {Data} & C@0.5IoU & B-4@0.5IoU & M@0.5IoU & R@0.5IoU & mAP@0.5IoU \\
 \hline
Scan2Cap & 3D & 32.94 & 20.63 & 21.10 & 41.58 & 27.45 \\
MORE & 3D & 38.98 & 23.01 & 21.65 & 44.33 & 31.93 \\
X-Trans2Cap & 3D & 41.52 & 23.83 & 21.90 & 44.97 & 34.68 \\
SpaCap3D & 3D & 42.53 & 25.02 & 22.22 & 45.65 & 34.44 \\
Ours & 3D & 42.77 & 23.60 & 22.05 & 45.13 & \textbf{36.99} \\
3DJCG & 3D & 47.68 & \textbf{31.53} & \textbf{24.28} & \textbf{51.08} & - \\
\hline
Ours\textsuperscript{*} & 3D & \textbf{50.29} & 25.64 & 22.57 & 44.71 & 35.97 \\
\hline
\hline
Scan2Cap & 2D+3D & 39.08 & 23.32 & 21.97 & 44.78 & 32.21 \\
D3Net & 2D+3D & 39.08 & 23.32 & 21.97 & 44.78 & 36.13 \\
MORE & 2D+3D & 40.94 & 22.93 & 21.66 & 44.42 & 33.75 \\
X-Trans2Cap & 2D+3D & 43.87 & 25.05 & 22.46 & 45.28 & 35.31 \\
SpaCap3D & 2D+3D & 44.02 & 25.26 & 22.33 & 45.36 & 36.64 \\
Ours & 2D+3D & 46.11 & 25.47 & 22.64 & 45.96 & \textbf{42.91} \\
3DJCG & 2D+3D & 49.48 & \textbf{31.03} & \textbf{24.22} & \textbf{50.80} & - \\
\hline
D3Net\textsuperscript{*} & 2D+3D & 47.32 & 24.76 & 21.66 & 43.62 & 38.03 \\
Ours\textsuperscript{*} & 2D+3D & \textbf{54.30} & 27.24 & 23.30 & 45.81 & 42.77 \\
\hline
\hline
\end{tabular}

\caption{Performance comparisons of  3D dense captioning results obtained by our and the state-of-the-art methods using VoteNet as the detector on the Nr3D dataset. ``-'' indicats that the results have not been disclosed. ``*'' indicates training with the CiDEr score reward.}
\vspace{+0.4cm}
\label{tab:2}
\begin{tabular}{l | c | c c c c c}
\hline
{Method} & {Data} & C@0.5IoU & B-4@0.5IoU & M@0.5IoU & R@0.5IoU & mAP@0.5IoU \\
 \hline
X-Trans2Cap & 3D & 30.96 & 18.70 & 22.15 & 49.92 & 34.13 \\
SpaCap3D & 3D & 31.43 & 18.98 & 22.24 & 49.79 & 33.17 \\
Ours & 3D & 34.67 & 20.22 & 22.54 & 50.88 & 38.12 \\
\hline
Ours\textsuperscript{*} & 3D & \textbf{35.86} & \textbf{20.73} & \textbf{22.86} & \textbf{51.23} & \textbf{38.35}\\
\hline
\hline
Scan2Cap & 2D+3D & 24.10 & 15.01 & 21.01 & 47.95 & 32.21 \\
X-Trans2Cap & 2D+3D & 33.62 & 19.29 & 22.27 & 50.00 & 34.38 \\
SpaCap3D & 2D+3D & 33.71 & 19.92 & 22.61 & 50.50 & 38.11 \\
Ours & 2D+3D & 35.26 & 20.42 & 22.77 & 50.78 & 39.29 \\
\hline
Ours\textsuperscript{*} & 2D+3D & \textbf{37.37} & \textbf{20.96} & \textbf{22.89} & \textbf{51.11} & \textbf{39.94} \\
\hline
\hline
\end{tabular}
\vspace{-0.2cm}
\end{table*}

\begin{figure*}
\centering
\includegraphics[width=\linewidth]{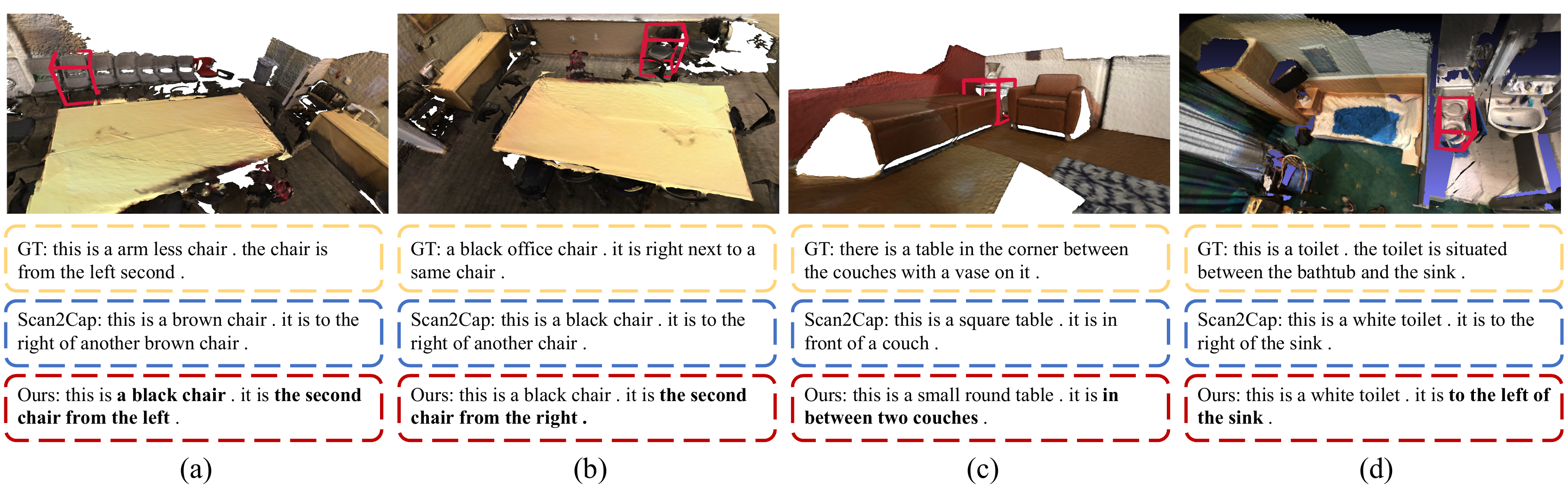} 
\vspace{-0.4cm}
\caption{Qualitative results of Scan2Cap and our proposed method. We only highlight the ground truth bounding box (in red) to locate the target object. 
It can be observed that our method performs superiorly to Scan2Cap in mining spatial relations between objects and generates more detailed and vivid descriptions closer to the  human annotations. 
Best viewed in color.}
\label{fig4}
\end{figure*}

\begin{table*}
\centering
\caption{Ablation studies on different components of our proposed model.}
\label{tab:3}
\resizebox{\linewidth}{!}{
\begin{tabular}{c | c c c c | c c c c c}
\hline
Model & GCM(object) & GCM(superpoint) & LCM(object) & LCM(superpoint) & C@0.5IoU & B-4@0.5IoU & M@0.5IoU & R@0.5IoU \\ 
\hline
A & \checkmark & & & & 43.91 & 24.63 & 22.22 & 45.54 \\ 
B & \checkmark & \checkmark & & & 44.71 & 25.20 & 22.37 & \textbf{46.09} \\ 
C & \checkmark & \checkmark & \checkmark &  & 45.93 & 25.14 & 22.48 & 45.53 \\ 
D & \checkmark & \checkmark & \checkmark & \checkmark & \textbf{46.11} & \textbf{25.47} & \textbf{22.64} & 45.96 \\ 
\hline
\end{tabular}
}
\vspace{-0.2cm}
\end{table*}

\subsection{Experimental Settings}
\textbf{Dataset and Metrics.}
Following the existing methods~\cite{chen2021scan2cap, yuan2022x, wang2022spatiality}, we evaluate our proposed method on the ScanRefer~\cite{chen2020scanrefer} and Nr3D~\cite{achlioptas2020referit3d} datasets. 
The ScanRefer dataset annotates 46,173 descriptions for 9,943 objects in 703 ScanNet 3D scenes, while the Nr3D dataset annotates 41,503 descriptions for 5,878 objects in 641 ScanNet 3D scenes. 
We follow the official ScanRefer benchmark splits to divide our data into train/validation/test sets containing 36,665, 9,508, and 5,410 samples, respectively. 
And, we follow the official Nr3D benchmark splits to divide our data for training and validation. Same as the previous work~\cite{chen2021scan2cap}, we perform the experiment comparisons and analyses on the validation set. 

To fully illustrate the effectiveness of our method, we evaluate the performances from the perspectives of both detection and captioning. 
For detection, we utilize the mean average precision (mAP) thresholded by 0.5 IoU score (mAP@0.5IoU) to evaluate our model. 
For captioning, 
we use multiple metrics $m$@0.5IoU defined in Scan2Cap~\cite{chen2021scan2cap}.
The metric $m$@0.5IoU only considers the description of the candidate objects whose IoU score of bounding box and GT bounding box are larger than 0.5. 
The captioning metric $m$ represents standard evaluation criteria for image captioning, such as CiDEr(C@0.5IoU), BLEU-4(B-4@0.5IoU), METEOR(M@0.5IoU), and ROUGE-L(R@0.5IoU). 

\textbf{Implementation Details.}
For the sake of fairness, we adopt VoteNet~\cite{qi2019deep} as the detector same as the existing methods.
Following Scan2Cap~\cite{chen2021scan2cap}, we set the number of input point clouds to 40,000 and use the same data augmentation strategy.
Specifically, the number of candidate objects $N_{\text{obj}}$ is 256, and the number of superpoints $N_{\text{sp}}$ is set as 1,024. 
For the context identification module, according to the number of objects in human-annotated sentences, we set the numbers of neighboring objects  and neighboring superpoints as 5 and 10, respectively.
The caption decoder consisting of two decoder layers.

\subsection{Performance Comparisons}

As shown in Table~\ref{tab:1}, we compare our method with Scan2Cap~\cite{chen2021scan2cap}, MORE~\cite{jiao2022more}, X-Trans2Cap~\cite{yuan2022x}, SpaCap3D~\cite{wang2022spatiality}, as well as D3Net~\cite{chen2021d3net} and 3DJCG~\cite{cai20223djcg}. Please note that D3Net and 3DJCG perform dual learning for the 3D dense captioning task, which unifies the dense captioning and visual grounding tasks into one framework.
The performance results are directly copied from the original paper or supplementary materials, with ``-'' indicating that the results have not been disclosed.
For a fair comparison, all of the compared methods use VoteNet~\cite{qi2019deep} as the detector.

We compare the results with different data types and training strategies.
First, we compare the results with and without 2D projection data.
``3D'' refers to using only the 3D coordinate $(x,y,z)$ as the input feature, while ``2D+3D'' means that the 2D multi-view features from the pre-trained ENet~\cite{paszke2016enet} and the normal coordinate from the point clouds are taken as the extra features besides the 3D coordinate information.
Compared with the results based on 3D data, the results of ``2D+3D'' achieve remarkable improvements in most evaluation metrics.
It shows that adding auxiliary features to the point clouds can enrich their feature representation ability and effectively boost the qualities of the  generated descriptions.
In addition, most models in the table use the cross-entropy loss for training, while ``*'' indicates training with the CiDEr score reward, such as $\text{D3Net}^*$ and $\text{Our}^*$.
From the results of $\text{D3Net}^*$ and $\text{Our}^*$, 
it can be observed that the reinforcement learning algorithm, specifically the self-critical training strategy, can further improve performance in most evaluation metrics, especially on the C@0.5IoU metric. 


Among these methods, Scan2Cap uses a message passing network and a 2-layer gated recurrent unit (GRU)  to generate descriptions and obtains 39.08\% on C@0.5IoU. MORE proposes a multi-order graph, which can efficiently capture the spatial relations of objects and achieves better performance.
However, Scan2Cap and MORE adopt an RNN-based caption decoder, which restricts the performance of their long sequence relationship modeling.
X-Trans2Cap proposes a Transformer-based method and introduces additional 2D data as auxiliary features, which improves 2.93\% on C@0.5IoU than MORE.
Besides, SpaCap3D designs a spatiality-guided Transformer to further investigate the relative spatiality of objects and achieve better performances.
By unifying dense captioning and visual grounding tasks, D3Net and 3DJCG achieve great improvements compared with the methods with single dense captioning task.
It is worth noting that the performance of 3DJCG is better than D3Net. The main reason is that 3DJCG adopts the Transformer structure to capture the relationship between candidate objects and generate the corresponding descriptions, while D3Net still relies on one RNN-based decoder.
However, these methods only use object features to mine spatial relationships but ignore the contextual information within the point clouds, especially the non-object details and background information.
Our proposed method incorporates both object features and contextual information depicted by the superpoints, achieving the best performance with \textbf{54.30\%} C@0.5IoU on the ScanRefer dataset. We believe that the performances can be further improved  by unifying the visual grounding and dense captioning tasks as one whole framework same as 3DJCG, which will be investigated in our future work.

Moreover, we also evaluate our method against the state-of-the-art methods on the Nr3D dataset.
As shown in Table~\ref{tab:2}, the same conclusion can be drawn. The experimental results on these two datasets show that our method is superior to the existing methods, which indicates the superiority of our method on modeling the contextual information within the point clouds.

\subsection{Qualitative Results}

To further illustrate the effectiveness of the contextual information in 3D dense captioning, some visualization results are illustrated in Fig.~\ref{fig4}.
As shown in Fig.\ref{fig4}~(a), our method can 
describe the color attributes ``\texttt{\small{black}}'' of the target object.
In addition, our method can capture the complicated relationships between objects, such as  ``\texttt{\small{the second chair from the left}}'' in Fig.\ref{fig4}~(a) and ``\texttt{\small{to the left of the sink}}'' in Fig.\ref{fig4}~(d) .
Moreover, with multiple similar objects, the description generated by Scan2Cap appears to be ambiguous, such as ``\texttt{\small{the right of another chair}}'' shown in  Fig.\ref{fig4}~(b). However, our proposed method can more accurately describes the absolute order relationship ``\texttt{\small{the second chair from the right}}'' in the whole 3D scene. Fig.\ref{fig4}~(c) illustrates that  our method further provides the detailed and accurate number of couches adjacent to the target object (``\texttt{\small{in between two couches}}'').

\subsection{Ablation Studies}

To further verify the effectiveness of our proposed components, we conduct ablation studies as shown in Table \ref{tab:3}. 
Here, we design four experiments to investigate the role of all objects and superpoints, as well as neighboring objects and superpoints.
Compared to model A which only models the relationships of all objects by GCM, model B additionally uses all superpoints. 
Thus, model B obtains a higher C@0.5mAP score than model A, which verifies that our extracted superpoints can improve the generated descriptions effectively by incorporating global contextual information.
Besides, model C further explores the relationship between neighboring objects with the LCM, getting better performance than model A and B.
The outcome demonstrates that neighboring objects can provide move effectively spatial relationship of objects than only using global features.
By adding neighboring superpoints, model D is significantly better than the aforementioned models, which validates the complementary power of objects and superpoints. These experiments demonstrate that, compared with utilizing GCM or LCM alone, the combination of them can bring further improvements.

%% file: 5_coclusion.tex
\section{Conclusion}
In this work, we first introduce point clouds clustering features as contextual information into the 3D dense captioning task. 
Specifically, we propose a contextual modeling method, including GCM and LCM modules.
GCM  exploits the relationships among all the candidate objects and superpoints to obtain more complete scene information of the whole point clouds, while LCM  establishes the relationship between the target object and its neighboring objects and superpoints to attain more complicated object features to enrich the object characterization.
Experiments on the ScanRefer and Nr3D datasets demonstrate the effectiveness of our proposed method. In the future, we will expand our work into the broader field of 3D vision and language tasks, such as 3D visual grounding. With richer contextual information of the 3D scenes being introduced, the corresponding performances are expected to be further improved.

%% file: aaai22.bbl
\begin{thebibliography}{30}
\providecommand{\natexlab}[1]{#1}

\bibitem[{Achlioptas et~al.(2020)Achlioptas, Abdelreheem, Xia, Elhoseiny, and
  Guibas}]{achlioptas2020referit3d}
Achlioptas, P.; Abdelreheem, A.; Xia, F.; Elhoseiny, M.; and Guibas, L. 2020.
\newblock Referit3d: Neural listeners for fine-grained 3d object identification
  in real-world scenes.
\newblock In \emph{European Conference on Computer Vision}, 422--440. Springer.

\bibitem[{Anderson et~al.(2018)Anderson, He, Buehler, Teney, Johnson, Gould,
  and Zhang}]{anderson2018bottom}
Anderson, P.; He, X.; Buehler, C.; Teney, D.; Johnson, M.; Gould, S.; and
  Zhang, L. 2018.
\newblock Bottom-up and top-down attention for image captioning and visual
  question answering.
\newblock In \emph{Proceedings of the IEEE conference on computer vision and
  pattern recognition}, 6077--6086.

\bibitem[{Cai et~al.(2022)Cai, Zhao, Zhang, Sheng, and Xu}]{cai20223djcg}
Cai, D.; Zhao, L.; Zhang, J.; Sheng, L.; and Xu, D. 2022.
\newblock 3DJCG: A Unified Framework for Joint Dense Captioning and Visual
  Grounding on 3D Point Clouds.
\newblock In \emph{Proceedings of the IEEE/CVF Conference on Computer Vision
  and Pattern Recognition}, 16464--16473.

\bibitem[{Chen, Chang, and Nie{\ss}ner(2020)}]{chen2020scanrefer}
Chen, D.~Z.; Chang, A.~X.; and Nie{\ss}ner, M. 2020.
\newblock Scanrefer: 3d object localization in rgb-d scans using natural
  language.
\newblock In \emph{European Conference on Computer Vision}, 202--221. Springer.

\bibitem[{Chen et~al.(2021{\natexlab{a}})Chen, Wu, Nie{\ss}ner, and
  Chang}]{chen2021d3net}
Chen, D.~Z.; Wu, Q.; Nie{\ss}ner, M.; and Chang, A.~X. 2021{\natexlab{a}}.
\newblock D3Net: A Speaker-Listener Architecture for Semi-supervised Dense
  Captioning and Visual Grounding in RGB-D Scans.
\newblock \emph{arXiv preprint arXiv:2112.01551}.

\bibitem[{Chen et~al.(2021{\natexlab{b}})Chen, Gholami, Nie{\ss}ner, and
  Chang}]{chen2021scan2cap}
Chen, Z.; Gholami, A.; Nie{\ss}ner, M.; and Chang, A.~X. 2021{\natexlab{b}}.
\newblock Scan2cap: Context-aware dense captioning in rgb-d scans.
\newblock In \emph{Proceedings of the IEEE/CVF Conference on Computer Vision
  and Pattern Recognition}, 3193--3203.

\bibitem[{Datta et~al.(2019)Datta, Sikka, Roy, Ahuja, Parikh, and
  Divakaran}]{datta2019align2ground}
Datta, S.; Sikka, K.; Roy, A.; Ahuja, K.; Parikh, D.; and Divakaran, A. 2019.
\newblock Align2ground: Weakly supervised phrase grounding guided by
  image-caption alignment.
\newblock In \emph{Proceedings of the IEEE/CVF International Conference on
  Computer Vision}, 2601--2610.

\bibitem[{Deshpande et~al.(2019)Deshpande, Aneja, Wang, Schwing, and
  Forsyth}]{deshpande2019fast}
Deshpande, A.; Aneja, J.; Wang, L.; Schwing, A.~G.; and Forsyth, D. 2019.
\newblock Fast, diverse and accurate image captioning guided by part-of-speech.
\newblock In \emph{Proceedings of the IEEE/CVF Conference on Computer Vision
  and Pattern Recognition}, 10695--10704.

\bibitem[{Jiao et~al.(2022)Jiao, Chen, Jie, Chen, Ma, and Jiang}]{jiao2022more}
Jiao, Y.; Chen, S.; Jie, Z.; Chen, J.; Ma, L.; and Jiang, Y.-G. 2022.
\newblock MORE: Multi-Order RElation Mining for Dense Captioning in 3D Scenes.
\newblock \emph{arXiv preprint arXiv:2203.05203}.

\bibitem[{Johnson, Karpathy, and Fei-Fei(2016)}]{johnson2016densecap}
Johnson, J.; Karpathy, A.; and Fei-Fei, L. 2016.
\newblock Densecap: Fully convolutional localization networks for dense
  captioning.
\newblock In \emph{Proceedings of the IEEE conference on computer vision and
  pattern recognition}, 4565--4574.

\bibitem[{Kim et~al.(2019)Kim, Choi, Oh, and Kweon}]{kim2019dense}
Kim, D.-J.; Choi, J.; Oh, T.-H.; and Kweon, I.~S. 2019.
\newblock Dense relational captioning: Triple-stream networks for
  relationship-based captioning.
\newblock In \emph{Proceedings of the IEEE/CVF Conference on Computer Vision
  and Pattern Recognition}, 6271--6280.

\bibitem[{Li et~al.(2019)Li, Zhu, Liu, and Yang}]{li2019entangled}
Li, G.; Zhu, L.; Liu, P.; and Yang, Y. 2019.
\newblock Entangled transformer for image captioning.
\newblock In \emph{Proceedings of the IEEE/CVF international conference on
  computer vision}, 8928--8937.

\bibitem[{Liu et~al.(2021)Liu, Zhang, Cao, Hu, and Tong}]{liu2021group}
Liu, Z.; Zhang, Z.; Cao, Y.; Hu, H.; and Tong, X. 2021.
\newblock Group-free 3d object detection via transformers.
\newblock In \emph{Proceedings of the IEEE/CVF International Conference on
  Computer Vision}, 2949--2958.

\bibitem[{Lu et~al.(2017)Lu, Xiong, Parikh, and Socher}]{lu2017knowing}
Lu, J.; Xiong, C.; Parikh, D.; and Socher, R. 2017.
\newblock Knowing when to look: Adaptive attention via a visual sentinel for
  image captioning.
\newblock In \emph{Proceedings of the IEEE conference on computer vision and
  pattern recognition}, 375--383.

\bibitem[{Luo et~al.(2021)Luo, Ji, Sun, Cao, Wu, Huang, Lin, and
  Ji}]{luo2021dual}
Luo, Y.; Ji, J.; Sun, X.; Cao, L.; Wu, Y.; Huang, F.; Lin, C.-W.; and Ji, R.
  2021.
\newblock Dual-level collaborative transformer for image captioning.
\newblock In \emph{Proceedings of the AAAI Conference on Artificial
  Intelligence}, volume~35, 2286--2293.

\bibitem[{Pan et~al.(2020)Pan, Yao, Li, and Mei}]{pan2020x}
Pan, Y.; Yao, T.; Li, Y.; and Mei, T. 2020.
\newblock X-linear attention networks for image captioning.
\newblock In \emph{Proceedings of the IEEE/CVF conference on computer vision
  and pattern recognition}, 10971--10980.

\bibitem[{Paszke et~al.(2016)Paszke, Chaurasia, Kim, and
  Culurciello}]{paszke2016enet}
Paszke, A.; Chaurasia, A.; Kim, S.; and Culurciello, E. 2016.
\newblock Enet: A deep neural network architecture for real-time semantic
  segmentation.
\newblock \emph{arXiv preprint arXiv:1606.02147}.

\bibitem[{Qi et~al.(2019)Qi, Litany, He, and Guibas}]{qi2019deep}
Qi, C.~R.; Litany, O.; He, K.; and Guibas, L.~J. 2019.
\newblock Deep hough voting for 3d object detection in point clouds.
\newblock In \emph{proceedings of the IEEE/CVF International Conference on
  Computer Vision}, 9277--9286.

\bibitem[{Qi et~al.(2017)Qi, Yi, Su, and Guibas}]{qi2017pointnet++}
Qi, C.~R.; Yi, L.; Su, H.; and Guibas, L.~J. 2017.
\newblock Pointnet++: Deep hierarchical feature learning on point sets in a
  metric space.
\newblock \emph{Advances in neural information processing systems}, 30.

\bibitem[{Rennie et~al.(2017)Rennie, Marcheret, Mroueh, Ross, and
  Goel}]{rennie2017self}
Rennie, S.~J.; Marcheret, E.; Mroueh, Y.; Ross, J.; and Goel, V. 2017.
\newblock Self-critical sequence training for image captioning.
\newblock In \emph{Proceedings of the IEEE conference on computer vision and
  pattern recognition}, 7008--7024.

\bibitem[{Song et~al.(2019)Song, Wang, Chen, and Jiang}]{song2019much}
Song, X.; Wang, B.; Chen, G.; and Jiang, S. 2019.
\newblock MUCH: Mutual Coupling Enhancement of Scene Recognition and Dense
  Captioning.
\newblock In \emph{Proceedings of the 27th ACM International Conference on
  Multimedia}, 793--801.

\bibitem[{Vaswani et~al.(2017)Vaswani, Shazeer, Parmar, Uszkoreit, Jones,
  Gomez, Kaiser, and Polosukhin}]{vaswani2017attention}
Vaswani, A.; Shazeer, N.; Parmar, N.; Uszkoreit, J.; Jones, L.; Gomez, A.~N.;
  Kaiser, {\L}.; and Polosukhin, I. 2017.
\newblock Attention is all you need.
\newblock \emph{Advances in neural information processing systems}, 30.

\bibitem[{Wang et~al.(2022)Wang, Zhang, Yu, and Cai}]{wang2022spatiality}
Wang, H.; Zhang, C.; Yu, J.; and Cai, W. 2022.
\newblock Spatiality-guided Transformer for 3D Dense Captioning on Point
  Clouds.
\newblock \emph{arXiv preprint arXiv:2204.10688}.

\bibitem[{Wang, Chen, and Hu(2019)}]{wang2019hierarchical}
Wang, W.; Chen, Z.; and Hu, H. 2019.
\newblock Hierarchical attention network for image captioning.
\newblock In \emph{Proceedings of the AAAI Conference on Artificial
  Intelligence}, volume~33, 8957--8964.

\bibitem[{Xu et~al.(2015)Xu, Ba, Kiros, Cho, Courville, Salakhudinov, Zemel,
  and Bengio}]{xu2015show}
Xu, K.; Ba, J.; Kiros, R.; Cho, K.; Courville, A.; Salakhudinov, R.; Zemel, R.;
  and Bengio, Y. 2015.
\newblock Show, attend and tell: Neural image caption generation with visual
  attention.
\newblock In \emph{International conference on machine learning}, 2048--2057.
  PMLR.

\bibitem[{Yang et~al.(2017)Yang, Tang, Yang, and Li}]{yang2017dense}
Yang, L.; Tang, K.; Yang, J.; and Li, L.-J. 2017.
\newblock Dense captioning with joint inference and visual context.
\newblock In \emph{Proceedings of the IEEE conference on computer vision and
  pattern recognition}, 2193--2202.

\bibitem[{Yin et~al.(2019)Yin, Sheng, Liu, Yu, Wang, and Shao}]{yin2019context}
Yin, G.; Sheng, L.; Liu, B.; Yu, N.; Wang, X.; and Shao, J. 2019.
\newblock Context and attribute grounded dense captioning.
\newblock In \emph{Proceedings of the IEEE/CVF Conference on Computer Vision
  and Pattern Recognition}, 6241--6250.

\bibitem[{Yuan et~al.(2022)Yuan, Yan, Liao, Guo, Li, Cui, and Li}]{yuan2022x}
Yuan, Z.; Yan, X.; Liao, Y.; Guo, Y.; Li, G.; Cui, S.; and Li, Z. 2022.
\newblock X-trans2cap: Cross-modal knowledge transfer using transformer for 3d
  dense captioning.
\newblock In \emph{Proceedings of the IEEE/CVF Conference on Computer Vision
  and Pattern Recognition}, 8563--8573.

\bibitem[{Zhang et~al.(2021)Zhang, Sun, Luo, Ji, Zhou, Wu, Huang, and
  Ji}]{zhang2021rstnet}
Zhang, X.; Sun, X.; Luo, Y.; Ji, J.; Zhou, Y.; Wu, Y.; Huang, F.; and Ji, R.
  2021.
\newblock RSTNet: Captioning with adaptive attention on visual and non-visual
  words.
\newblock In \emph{Proceedings of the IEEE/CVF conference on computer vision
  and pattern recognition}, 15465--15474.

\bibitem[{Zhong et~al.(2020)Zhong, Wang, Chen, Yu, and
  Li}]{zhong2020comprehensive}
Zhong, Y.; Wang, L.; Chen, J.; Yu, D.; and Li, Y. 2020.
\newblock Comprehensive Image Captioning via Scene Graph Decomposition.
\newblock In \emph{Computer Vision--ECCV 2020: 16th European Conference,
  Glasgow, UK, August 23--28, 2020, Proceedings, Part XIV}, 211--229.

\end{thebibliography}
